\pdfoutput=1

\documentclass[11pt]{article}

\usepackage{emnlp2021}

\usepackage{times}
\usepackage{latexsym}

\usepackage[T1]{fontenc}

\usepackage[utf8]{inputenc}

\usepackage{microtype}

\usepackage[noend]{algpseudocode}
\usepackage{algorithmicx,algorithm}
\usepackage{amsmath}
\usepackage{bm}
\usepackage{amsfonts}
\usepackage{enumerate}
\usepackage{fancyhdr}
\usepackage{multirow,booktabs,makecell}
\usepackage{graphicx}

\title{A Syntax-Guided Grammatical Error Correction Model with Dependency Tree Correction}

\author{Zhaohong Wan\textsuperscript{\rm 1,\rm 2} \and Xiaojun Wan\textsuperscript{\rm 1,\rm 2} \\
	\textsuperscript{\rm 1}Wangxuan Institute of Computer Technology, Peking University \\
	\textsuperscript{\rm 2}The MOE Key Laboratory of Computational Linguistics, Peking University \\
	{\tt \{xmwzh,wanxiaojun\}@pku.edu.cn} \\}

\begin{document}
\maketitle
\begin{abstract}
Grammatical Error Correction (GEC) is a task of detecting and correcting grammatical errors in sentences. Recently, neural machine translation systems have become popular approaches for this task. However, these methods lack the use of syntactic knowledge which plays an important role in the correction of grammatical errors. In this work, we propose a syntax-guided GEC model (SG-GEC) which adopts the graph attention mechanism to utilize the syntactic knowledge of dependency trees. Considering the dependency trees of the grammatically incorrect source sentences might provide incorrect syntactic knowledge, we propose a dependency tree correction task to deal with it. Combining with data augmentation method, our model achieves strong performances without using any large pre-trained models. We evaluate our model on public benchmarks of GEC task and it achieves competitive results.
\end{abstract}

\section{Introduction}
Grammatical Error Correction (GEC) is a task of detecting and correcting grammatical errors in sentences.
Due to the growing number of language learners of English, there has been increasing attention to the English GEC in the past few years. 

Considering the outstanding performance of neural network models in machine translation tasks, numerous studies have applied cutting-edge neural machine translation models to GEC task\citep{Zhao2019ImprovingGE,JunczysDowmunt2018ApproachingNG} 
. Besides, the adoption of large pre-trained models becomes popular as well \citep{Kaneko2020EncoderDecoderMC,omelianchuk2020gector}. These works have achieved great success, but lack the use of syntactic knowledge which plays an important role in the correction of grammatical errors.

In this work, we propose a syntax-guided GEC model (SG-GEC) with dependency tree correction task to exploit the syntactic knowledge of dependency trees. A dependency tree is a directed graph representing syntactic knowledge of several words towards each other. Inspired by \citet{velivckovic2017graph}, we adopt the graph attention mechanism to utilize the syntactic knowledge within dependency trees. 

Especially, the source sentences in GEC task are sentences with grammatical errors which means the dependency trees of source sentences might provide incorrect syntactic knowledge. So simply applying the graph attention mechanism over source sentences would not work well. Given that, we proposed a dependency tree correction task to construct dependency trees of corrected sentences.
Considering a tree can be uniquely determined by relations of nodes, we construct the dependency trees of corrected sentences by predicting the relations of nodes instead of the entire tree. By applying this additional task, the model can construct dependency trees of corrected sentences and enrich the model with the corrected syntactic knowledge.

We apply the data augmentation method to further improve the performance of the model. Experiments are conducted on the following widely used benchmarks: CoNLL-2014 \citep{Ng2014The}, FCE \citep{Yannakoudakis2011AND}, BEA-2019 \citep{Bryant2019TheBS}.  Among models without using the large pre-trained models, our model achieves the best F-score on all benchmarks. Comparing with models which incorporate the large pre-trained models, our model achieves very competitive performance as well. 
In general, our model achieves strong performance without using any large pre-trained models.

Our contributions are summarized as follows:

1. To the best of our knowledge, we introduce syntactic knowledge into neural GEC model for the first time, by applying graph attention mechanism to utilize the dependency tree.

2. 
We propose a dependency tree correction task to deal with the problem that the dependency trees of grammatically incorrect source sentences might provide incorrect syntactic knowledge.

3. Without using any large pre-trained model, our SG-GEC model achieves strong performances on public GEC benchmarks.

\section{Related Work}

Early published works in GEC developed models based on manually designed grammar rules
\citep{Murata1994DeterminationOR,Bond1996NounPR,siegel1996preferences}.  After \citet{Han2006DetectingEI} pointed out the limitation of rule-based method, some researchers turned their attention to the statistical machine learning method \citep{Knight1994AutomatedPO,Minnen2000MemoryBasedLF,Izumi2003AutomaticED}. 

With the development of deep learning, recent works proposed various neural network models to solve GEC task. Some regarded the GEC task as a translation problem and applied cutting-edge neural machine translation model to deal with it \citep{ Yuan2016GrammaticalEC, Chollampatt2018AMC}.
Many recent works \citep{JunczysDowmunt2018ApproachingNG, Zhao2019ImprovingGE} made use of the powerful machine translation architecture Transformer \citep{Vaswani2017AttentionIA}. Considering the tremendous performance of pre-trained methods, pre-trained language model, such as BERT\citep{Devlin2019BERTPO},RoBERTa\citep{liu2019roberta} and XLNet\citep{yang2019xlnet}, have been adopted in GEC models\citep{Kaneko2020EncoderDecoderMC,omelianchuk2020gector}.

A challenge in applying neural machine translation models to GEC task is the requirement of the large training data. Given that, many works incorporated data augmentation methods to address this problem.
Many works adopted pre-defined rules to generate synthetic samples with grammatical errors. \citep{Grundkiewicz2019NeuralGE,Lichtarge2018WeaklySG,Choe2019ANG}. 
\citet{Kiyono2019AnES} further studied the data augmentation methods and showed the efficacy of back-translation procedure.

Recently, dependency parsing has been further developed with neural network\citep{dozat2016deep,li2018seq2seq}. Benefiting from it, models could receive syntactic knowledge with higher accuracy. Many works showed the potential of using syntactic knowledge in various  tasks\citep{zhang2020sg,wang2020better,jin2020semsum}. Inspired by previous works, we proposed the SG-GEC model to utilize the syntactic knowledge within dependency trees.

\section{Our Approach}

\begin{figure*}
  \centering
  \includegraphics[width=0.8\textwidth]{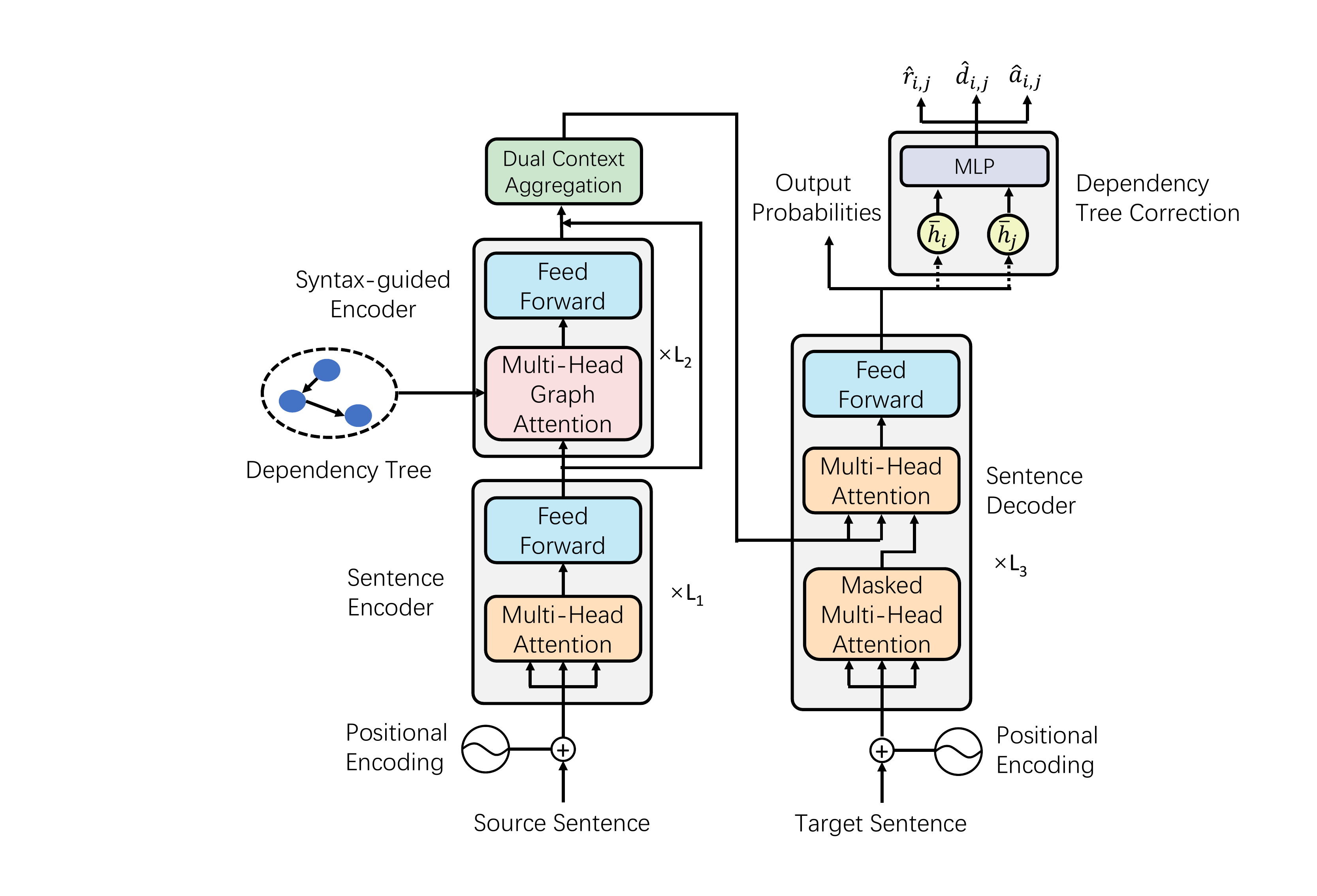} 
  \caption{Overview of SG-GEC model} 
  \label{fig.model} 
\end{figure*}

In this section, we will introduce our proposed approach.
Firstly, we begin by providing the notations we use.
Then we introduce the syntax-guided model which consists of a sentence encoder, a syntax-guided encoder and a sentence decoder. Next, we show the proposed dependency tree correction task. Finally, the objective functions will be presented. 
The overall architecture is shown in Figure \ref{fig.model}.

\subsection{Notations}

Given GEC training data $\mathcal{D}$ that comprise pairs of grammatically incorrect source sentence $X=(x_{1},x_{2},...,x_{N})$ and grammatically correct target sentence $Y=(y_{1},y_{2},...,y_{M})$, where $N$ is the length of source sentence and $M$ is the length of target sentence.  

With a dependency parser, we can get the dependency trees $G_{x}=(V_{x},E_{x})$ of the source sentence $X$ and $G_{y}=(V_{y},E_{y})$ of the target sentence $Y$, where $V_{x}$, $V_{y}$ are sets of nodes, $E_{x}$, $E_{y}$ are sets of edges, respectively. Each node represents a token in the sentence. We denote $R$ as the set of dependency relation labels. Each edge of dependency tree is denoted as a triple $(i,j,r_{i,j})$, which means a dependency relation from the node $v_{i}$  to the node $v_{j}$ with the relation label $r_{i,j} \in R$.

\subsection{Sentence Encoder}

Our model adopts Transformer architecture as sentence encoder to construct the context-aware representations of source sentence. The sentence encoder encodes the source sentence $X=(x_{1},x_{2},...,x_{N})$  with a stack of $L_{1}$ identical layers which applies multi-head attention mechanism to tokens followed by feed-forward layers.

Let $H^l=(h_{1}^l,h_{2}^l,...,h_{N}^l)$ denote the representations of tokens at layer $l$. In particular, the input of sentence encoder  $H^0=(h_{1}^0,h_{2}^0,...,h_{N}^0)$ is the sum of the token embedding and position encoding.

It performs the multi-head attention to get the values 
and adopts residual connection \citep{He2016IdentityMI} and layer normalization \citep{ba2016layer} to connect adjacent layers.
The final contextual representation is denoted as $H^{L_{1}}$.

\subsection{Syntax-Guided Encoder}

In order to incorporate the syntactic knowledge of dependency tree, we proposed a syntax-guided encoder. It is a stack of
$L_{2}$ identical graph layers which consist of graph attention mechanism and fully connected feed-forward network. 

Let $\hat{H}^l=(\hat{h}_{1}^l,\hat{h}_{2}^l,...,\hat{h}_{N}^l)$ denote the representation of tokens at layer $l$. In particular, the input of syntax-guided encoder
is the output of sentence encoder $H^{L_{1}}$.

Inspired by \citet{velivckovic2017graph}, we adopt the graph attention mechanism to exploit the syntactic knowledge within the dependency tree $G_{x}$ of source sentence $X$. However, this proposed graph attention mechanism is designed for undirected graphs without labels, while the dependency tree is a directed graph with relation labels. So we treat the dependency relations as the nodes as well, and get the representation of each node by aggregating the neighbor relations.

Firstly, we get the representation of a relation $(i,j,r_{i,j})$ from the representations of nodes $v_{i}$ and $v_{j}$, and the learnable embedding of relation label $r_{i,j}$.
Considering incoming edges and outgoing edges might play different roles to a node, we apply different mappings over incoming relations and outgoing relations, respectively.
The representation of outgoing relation $(i, j, r_{i,j})$ for node $v_{i}$ is computed
as follows:

\begin{equation}
\overrightarrow{u_{i,j}}=\operatorname{ReLU}\left(\left[\hat{h}_{i}^l\left\|e_{i,j}^r\right\| \hat{h}_{j}^l\right] W_{\text {out }}+b_{\text {out }}\right)\end{equation}

\noindent where $e^r_{i,j} \in \mathbb{R}^{d_{\text {model }}} $ is the
embedding of relation label $r_{i,j}$ , $W_{out} \in \mathbb{R}^{d_{\text {model }}\times 3d_{\text {model }}}$ 
and $b_{out} \in \mathbb{R}^{d_{\text {model }}}$ are the parameters.

Similarly, the
representation of incoming relation $(j, i, r_{j,i})$ for node $v_{i}$ is computed as
follows:

\begin{equation}
\overleftarrow{u_{j,i}}=\operatorname{ReLU}\left(\left[\hat{h}_{j}^l\left\|e_{j,i}^r\right\| \hat{h}_{i}^l\right] W_{\text {in }}+b_{\text {in }}\right)\end{equation}

\noindent where  $e^r_{j,i} \in \mathbb{R}^{d_{\text {model }}} $ is the
embedding of relation label $r_{j,i}$, $W_{in} \in \mathbb{R}^{d_{\text {model }}\times 3d_{\text {model }}}$ 
and $b_{in} \in \mathbb{R}^{d_{\text {model }}}$.

With graph attention mechanism, we can get a new representation $\hat{h}_{i}^{*}$ of node $v_{i}$, by aggregating the representation of its neighbor outgoing relations  and neighbor incoming  relations. This process can be formulated as follows:

\begin{equation}
\hat{h}_{i}^* =\sum_{u \in \mathcal{NR}_{x}(v_{i})} \alpha\left(u, \hat{h}_{i}\right) u W^{V} 
\end{equation}

\begin{equation}
\alpha\left(u, \hat{h}_{i}\right) =\frac{\exp \left(\left(u W^{K}\right)^{T} \hat{h}_{i} W^{Q}\right)}{\sum_{z \in \mathcal{NR}_{x}(v_{i})} \exp \left(\left(z W^{K}\right)^{T} \hat{h}_{i} W^{Q}\right)}
\end{equation}

\noindent where $\mathcal{NR}_{x}(v_{i})$ denotes all neighbor relations of $v_{i}$ in dependency tree of source sentence $G_{x}$, including incoming relations and outgoing relations.

Multi-head operation was adopted in graph attention as well: 

\begin{equation}
    \mathrm{MHGAT}\left(\hat{h}_{i}\right)=\left(\|_{t=1}^{T} \hat{h}_{i}^{*t}\right) W^{O}    
\end{equation}

\noindent where $\|$ denotes the concatenation of the $T$ attention heads and
$\hat{h}_{i}^{*t}$ is the result $\hat{h}_{i}^{*}$ of graph attention in head $t$. Each head $t$ learns independent transformations $W_{t}^{Q}, W_{t}^{K} \in \mathbb{R}^{d_{\text {model }} \times d_{k}}$,
$W_{t}^{V} \in \mathbb{R}^{d_{\text {model }} \times d_{v}}$, $W^{O} \in \mathbb{R}^{Td_{v} \times d_{\text {model }}} 
$
respectively.

The fully connected feed-forward network adopts residual connection and layer normalization
for connecting the adjacent layers.

\begin{equation}
\hat{H}^\prime =\text { LayerNorm }\left(\hat{H}^{l-1}+\mathrm{MHGAT}\left(\hat{H}^{l-1}\right)\right) 
\end{equation}

\begin{equation}
\hat{H}^{l} =\text { LayerNorm }(\hat{H}^\prime+\mathrm{FFN}(\hat{H}^\prime))
\end{equation}

\noindent where $l \in\left[1, L_{2}\right]$, and the final representation of syntax-guided encoder is $\hat{H}^{L_{2}}$.

\subsection{Dual Context Aggregation}
Considering that we have two
representations: one is the contextual representation $H^{L_1}=(h_{1}^{L_1},h_{2}^{L_1},...,h_{N}^{L_1})$ from the
sentence encoder, the other is syntax-guided representation $\hat{H}^{L_2}=(\hat{h}_{1}^{L_2},\hat{h}_{2}^{L_2},...,\hat{h}_{N}^{L_2})$
from syntax-guided encoder. Formally, the
output $O=(o_{1},o_{2},...,o_{N})$ of the whole encoder is computed by:
\begin{equation}
    o_{i}=\beta h_{i}^{L_1}+(1-\beta)\hat{h}_{i}^{L_2}
\end{equation}
where $\beta \in (0,1)$ is a hyper-parameter.  

\subsection{Sentence Decoder}

The sentence decoder has an architecture similar
to Transformer. It is composed of $L_{3}$ identical
layers. Each layer has three sub-layers: a masked multi-head self-attention
mechanism, a multi-head attention mechanism and a feed-forward network.

Let $\overline{H}^l$ denotes the representations of tokens at layer $l$. Similar to the sentence encoder, the input $\overline{H}^0$ of sentence decoder is the sum of positional encoding and token embedding.

Firstly, we use the masked multi-head self-attention mechanism to encode the sequences.
Then, the encoded sequence and the output of encoder $O$ are fed to the multi-head attention mechanism and feed-forward network. 

For convenience, we denote the final output of the
decoder $\overline{H}^{L_{3}}$ as $\overline{H}$. It is passed through a softmax layer to generate the probability distribution
$p_{i}^g$ of the next word over the target vocabulary at step $i$.

Considering the similarity between input and output, we adopt copy mechanism to
improve the performance of our model, which can deal with the problem of out-of-vocabulary words as well.

At each generation step $i$, we obtain copy distribution $p_{i}^{c}$ from the output of decoder $\overline{H}$ and the output of encoder $O$. Then, we calculate the generation probability $\eta_{i} \in[0,1]$ from the  output of decoder $\overline{H}$. 

The final distribution $p_{i}$ at step $i$ is the weighted average of the two probability distributions:
\begin{equation}
p_{i}=\eta_{i} * p_{i}^{g}+\left(1-\eta_{i}\right) * p_{i}^{c}
\end{equation}

\subsection{Dependency Tree Correction}

Unlike other tasks, the source sentences in GEC task are sentences with grammatical errors. It means the dependency trees of source sentences might have errors as well. So only applying the graph attention mechanism over source sentences would get the syntactic knowledge with errors. To remedy this, we propose an auxiliary  dependency tree correction task to construct dependency trees of corrected sentences and enrich the model with the corrected syntactic knowledge. 

As for any two nodes in a dependency tree, there are three groups of relations between them:
dependency relation, distance and ancestor-descendant relation. The tree can be uniquely determined by these relations of nodes. So we can construct the dependency trees of corrected sentences by using the final representations of the decoder to predict these relations of nodes. 

The proposed dependency tree correction task consists of three sub-tasks. Each sub-task corresponds to one of the three relations between two nodes.

The first sub-task requires the model to predict the dependency relations for the
given node pairs. Given the outputs $\overline{h}_{i} \in \overline{H}$, $\overline{h}_{j} \in \overline{H}$ of the decoder for two nodes $v_i$ and $v_j$, we adopt a
multi-layer perceptron to predict the corresponding dependency
relation.

\begin{equation}
\hat{r}_{i, j}=\operatorname{softmax}\left(W_{r}\left[\operatorname{MLP}\left(\left[\overline{h}_{i} \|\overline{h}_{j}\right]\right)\right]+b_{r}\right)
\end{equation}

\noindent where $W_{r} \in \mathbb{R}^{(L+1) \times d_{\text {model }}}, b_{r} \in \mathbb{R}^{L+1}$, and $L$ is the number of dependency relation label types in the dependency tree. 

For a pair of nodes that are adjacent in the dependency tree, the gold label is the given dependency relation $r_{i, j}$. For a pair of nodes that are not adjacent, the gold label is non-adjacent.

The second sub-task requires the model to predict the distance between a pair of nodes in the dependency tree. The distance
$d_{i,j}$ of a pair of nodes is defined as the length of the shortest
path from $v_{i}$ to $v_{j}$ regardless of the direction. 

The distance of two nodes is a significant syntactic knowledge in dependency tree. It helps the model to capture the non-local knowledge.  Distance is predicted as follows:

\begin{equation}
\hat{d}_{i, j}=\operatorname{softmax}\left(W_{d}\left[\operatorname{MLP}\left(\left[\overline{h}_{i} \| \overline{h}_{j}\right]\right)\right]+b_{d}\right)
\end{equation}

\noindent where $W_{d} \in \mathbb{R}^{(D+1) \times d_{\text {model }}}, b_{d} \in \mathbb{R}^{D+1}$, and $D$ is the maximum distance of the dependency graphs in the dataset. $d_{i, j}$ is the ground truth.

The last sub-task requires the model to predict the ancestor-descendant relation between a pair of nodes. Given a node $v$, if any node $v^\prime$ is on the path from root to node $v$, then node $v^\prime$ is the ancestor of node $v$ and node $v$ is the descendant of node $v^\prime$. We denote $a_{i,j}$ as the ancestor-descendant relation from node $v_{i}$ to node $v_{j}$. If node $v_{i}$ is the ancestor of node $v_{j}$,  the ancestor-descendant relation $a_{i,j}$ is ancestor. If node $v_{j}$ is the ancestor of node $v_{i}$,  the ancestor-descendant relation $a_{i,j}$ is descendant. Otherwise, there is no ancestor-descendant relation between two nodes.

It is an important syntactic knowledge and we predict the relation as follows:

\begin{equation}
\hat{a}_{i, j}=\operatorname{softmax}\left(W_{a}\left[\operatorname{MLP}\left(\left[\overline{h}_{i} \| \overline{h}_{j}\right]\right)\right]+b_{a}\right)
\end{equation}

\noindent where $W_{a} \in \mathbb{R}^{3 \times d_{\text {model }}}, b_{a} \in \mathbb{R}^3$. For a pair of nodes on the same path from root to a leaf node, the ground truth is the ancestor-descendant relation $a_{i,j}$. Otherwise, the ground truth is non-relation.

\subsection{Objective Function}

Given a source sentence $X$ and the corresponding dependency tree $G_{x}$, the grammatical error correction objective is to maximize the probability of grammatically correct target sentence $Y$. The following negative log-likelihood function is optimized:

\begin{equation}
\mathcal{L}_{g}=-\sum_{i=1}^{M} \log P\left(y_{i} \mid y_{1: i-1}, X,G_{x}, \theta\right)
\end{equation}

\noindent where $\theta$ represents the  parameters of model.

To construct the dependency tree of corrected sentence and enrich the model with the corrected syntactic knowledge. We also optimize the three proposed dependency tree correction objectives. Considering the inconsistency between the tokens of generated sentence and the target sentence, we only compute the loss function on the overlapped nodes. For
 convenience, we denote $S(\hat{Y},Y)$ as the set of pairs of overlapped nodes.

The dependency relation prediction objective is defined as follows:

\begin{equation}
\mathcal{L}_{r}=-\frac{M}{|S|} \sum_{(i,j)\in S(\hat{Y},Y)}^{|S|}  \log P\left(r_{i, j} \mid X,Y,G_{x}, \theta\right)
\end{equation}

\noindent where $|S|$ is the size of $S(\hat{Y},Y)$, $r_{i, j}$ is the ground truth for the relation of nodes $v_{i}$ and $v_{j}$ in dependency tree $G_{y}$.

The distance prediction objective is defined as follows:
\begin{equation}
\mathcal{L}_{d}=-\frac{M}{|S|} \sum_{(i,j)\in S(\hat{Y},Y)}^{|S|} \log P\left(d_{i, j} \mid X,Y,G_{x}, \theta\right)
\end{equation}

\noindent where $d_{i, j}$ is the ground truth for the distance of nodes $v_{i}$ and $v_{j}$ in dependency tree $G_{y}$.

The ancestor-descendant relation prediction objective is defined as follows:
\begin{equation}
\mathcal{L}_{a}=-\frac{M}{|S|} \sum_{(i,j)\in S(\hat{Y},Y)}^{|S|} \log P\left(a_{i, j} \mid X,Y,G_{x}, \theta\right)
\end{equation}

\noindent where $a_{i, j}$ is the ground truth for the ancestor-descendant relation of nodes $v_{i}$ and $v_{j}$ in dependency tree $G_{y}$.

Thus, the overall loss  is the weighted sum of the grammatical error correction objective and dependency tree correction objectives:

\begin{equation}
\mathcal{L}=\mathcal{L}_{g}+\lambda_{1} * \mathcal{L}_{r}+\lambda_{2} * \mathcal{L}_{d}+\lambda_{3} * \mathcal{L}_{a}
\end{equation}

\noindent where $\lambda_{1}$ , $\lambda_{2}$ and $\lambda_{3}$ are the hyper-parameters.

\section{Experiment Setup}

\subsection{Datasets}

We use the following GEC datasets as original training corpus: National University
of Singapore Corpus of Learner English (NUCLE)\cite{Dahlmeier2013BuildingAL}, Lang-8 Corpus of Learner English (Lang-8)\cite{Tajiri2012TenseAA},
FCE dataset\cite{Yannakoudakis2011AND}, and Write \& Improve + LOCNESS Corpus(W\&I+LOCNESS)\cite{Bryant2019TheBS}.

We report results on CoNLL-2014 benchmark evaluated by official M2 scorer\footnote{https://github.com/nusnlp/m2scorer}\cite{Dahlmeier2012BetterEF}, and
on BEA-2019 and FCE benchmarks evaluated by ERRANT\footnote{https://github.com/chrisjbryant/errant}.

\subsection{Model Details}

\begin{table*}[t]
\centering
\footnotesize
\begin{tabular}{|l c c p{0.45cm} p{0.45cm} p{0.45cm} p{0.45cm} p{0.45cm} p{0.45cm} p{0.45cm} p{0.45cm} p{0.45cm}|}
\hline
\multirow{2}{*}{\bf Method}&\multirow{2}{1.6cm}{\bf Pre-trained Model}&\multirow{2}{1.5cm}{\bf Augmented  Data Size}&
\multicolumn{3}{c}{\bf CoNLL-2014}&\multicolumn{3}{c}{ \bf BEA-2019}&\multicolumn{3}{c|}{\bf FCE-test}\cr 
\cline{4-12}
    &&&P&R&$F_{0.5}$&P&R&$F_{0.5}$&P&R&$F_{0.5}$\cr
\hline
\bf Single Model&&&&&&&&&&&\cr
\hline
\citet{lichtarge2019corpora}&-&170M&65.5&37.1&56.8&-&-&-&-&-&-\cr
\citet{Zhao2019ImprovingGE}&-&30M&67.7&40.6&59.8&-&-&-&-&-&-\cr
\citet{Kiyono2019AnES}&-&70M&67.9&44.1&61.3&65.5&59.4&64.2&-&-&-\cr
SG-GEC(ours) &-&10M&74.4&39.5&63.2 &74.5&{48.6}& 67.3& {\bf 67.8}&{ 43.1}& {\bf 60.8}\cr
\hline
\citet{Kaneko2020EncoderDecoderMC}&\checkmark&70M&69.2 &{\bf45.6}& 62.6&67.1 &{\bf60.1}& 65.6&  59.8 &{\bf46.9}& 56.7\cr
\citet{omelianchuk2020gector} &\checkmark&9M&{\bf77.5}&40.1&{\bf65.3}&{\bf79.2}&53.9&{\bf72.4}&-&-&-\cr
\hline
\bf Ensemble Model&&&&&&&&&&&\cr
\hline
\citet{lichtarge2019corpora}&-&170M&66.7&43.9&60.4&-&-&-&-&-&-\cr
\citet{Zhao2019ImprovingGE}&-&30M&71.6&38.7&61.2&-&-&-&-&-&-\cr
\citet{Kiyono2019AnES}&-&70M&73.3&44.2&64.7&74.7&56.7&70.2&-&-&-\cr
SG-GEC(ours) &-&10M&{\bf78.7}&{41.7}&{\bf66.8}&{\bf80.9}&{49.3}&71.7&{\bf71.4}&{ 43.0}& {\bf 63.1}\cr
\hline
\citet{Kaneko2020EncoderDecoderMC}&\checkmark&70M&{72.6}& {\bf46.4}&65.2&72.3& \bf 61.4& 69.8&  62.8& \bf 48.8& 59.4 \cr
\citet{omelianchuk2020gector} &\checkmark&9M&78.2&41.5&66.5&78.9&58.2&\bf 73.6&-&-&-\cr
\hline
\end{tabular}
\caption{Comparison results of GEC methods. The top group shows the results of the single models. The second group shows the results of the ensemble models. Pre-trained model column shows whether the method adopts any large pre-trained models.
Augmented data sizes show the amounts of additional training sentences used in each method.  \textbf{Bold} indicates the highest score in each column. 
  }
\label{tab.topmodel}
\end{table*}

In this paper, we use the public toolkit Fairseq \cite{ott2019fairseq} to build our model.

We set 4 heads for multi-head graph attention and 8 heads for multi-head self-attention, masked multi-head self-attention. The dimension of word embedding and hidden units is 512.
For the inner layer in the feed-forward network, the size is 2048. The numbers 
of layers $L_{1}$, $L_{2}$, $L_{3}$ are set to 6, 3 and 6, respectively. The weight $\beta$ in the dual context aggregation is 0.5. The weight coefficients of loss $\lambda_{1}$, $\lambda_{2}$, $\lambda_{3}$ are set to 0.5, 0.1, 0.1.

To avoid unknown tokens in datasets, we apply byte-pair encoding (BPE) \cite{Sennrich2016NeuralMT} to sentences before feeding the texts into models. This operation may divide a word into several sub-words, which might lead to the mismatch between the nodes in the dependency tree and the tokens in the source sequence. Considering that, we take the average of the word embeddings of all sub-words corresponding to a word as the input representation of this word.

The model is trained by using the Adam optimization method \cite{Kingma2015AdamAM} with  initially learning rate 0.0001, momentum $\beta_{1}$ = 0.9, $\beta_{2}$ = 0.999
and weight decay $10^{-5}$. To avoid overfitting, we adopt dropout mechanism \cite{Srivastava2014DropoutAS} with dropout rate 0.1. Beam search with beam size of 5 is used for decoding.

\subsection{Training Stages}

Following \citet{Omelianchuk2020GECToRG}, we train the GEC model in three stages.

Firstly, we train the model on synthetic sentences generated by data augmentation method. Then, we extract sentence pairs containing grammatical errors from four training datasets (NUCLE, Lang-8, FCE and W\&I+LOCNESS) and fine-tune the model on these sentence pairs. Finally, we fine-tune the model on the respective entire training dataset corresponding to each test set.

\subsection{Implementation Details}
The incorporation of data augmentation method has been one of the most effective ways to improve the performance of GEC models. Following previous work, we use 10M parallel sentences with synthetically generated grammatical errors \citep{Awasthi2019ParallelIE}.

We parse the  sentences of these datasets with a dependency parser\citep{dozat2016deep}, which is a neural graph-based model 
and achieves very competitive results on standard treebanks for six different languages.

To further improve the performance, we incorporate the following techniques that are widely used in GEC task. 
Following \citet{Choe2019ANG}, we use spellcheck Enchant\footnote{https://github.com/pyenchant/pyenchant} to assist the correction of spelling errors. 
Following \citet{Sennrich2016EdinburghNM,Sennrich2017TheUO}, we use the right-to-left re-ranking(R2L) method to build the ensemble of independently trained models. We pass n-best candidates generated from four left-to-right models to  four right-to-left models, and re-rank the n-best candidates based on their corresponding scores.

\section{Results and Analysis}

\subsection{Comparison Results}

We evaluate the performance of our SG-GEC model on public benchmarks and compare the scores with the current top models in GEC task. Table \ref{tab.topmodel} shows the results. 

Comparing with previous models without using the large pre-trained models, our SG-GEC achieves the best F-score on all benchmarks.  It outperforms not only all previous single models but also all ensemble models with less augmented data.

Comparing with models which incorporate the large pre-trained models, our model achieves very competitive performance as well. As for single models, our model outperforms the method that adopts pre-trained model BERT and large amounts of augmented data\cite{Kaneko2020EncoderDecoderMC}. As for ensemble models, our model achieves the best F-score on CoNLL-2014 benchmark and FCE benchmark. Besides, it achieves very strong performance on BEA-2019 benchmark without using any large pre-trained models. Comparing with the method \cite{omelianchuk2020gector} that adopts three large pre-trained models which leads to huge computing costs, the results of our ensemble model are competitive as well.

As we can see from the table, our ensemble model achieves the highest precision in all benchmarks and our single model achieves high precision as well.
It indicates that the syntactic knowledge of sentences captured by our proposed SG-GEC model can improve the performance of the model.

By incorporating the syntactic knowledge, our SG-GEC model achieves strong performances on three public benchmarks without using any large pre-trained models.

\begin{table}[h]
\begin{center}
\footnotesize
\begin{tabular}{|p{3.5cm}| c c l|}
    \hline
    \multirow{2}{*}{\bf Model}&
    \multicolumn{3}{c|}{\bf CoNLL-2014}\cr
    \cline{2-4}
    &P&R&$F_{0.5}$\cr
    \hline
    Copy-Transformer& 71.8&39.9&61.9\cr
    \hline
    $\text{Copy-Transformer}$  $+\text{Dependency tree correction}$& 73.4&41.2&$63.5^{*}$\cr
     $\text{Copy-Transformer}$ $+\text{Syntax-guided encoder}$&77.0&40.3&$65.1^{***}$ \cr
    SG-GEC&78.7&41.7&$66.8^{***}$\cr
    
    \hline
    \end{tabular}
    \end{center}
    \caption{ Results of ablation study on CoNLL-2014 benchmark. For the last three models, we do the significance tests of F-score between these model and Copy-Transformer respectively. $*$ means the p-value < 0.1. $**$ means the p-value < 0.05. $***$ means the p-value < 0.01 }
    \label{tab.abl}
\end{table}

\subsection{Ablation Study}

\begin{table*}[h]
\begin{center}
\begin{tabular}{|l| c c l|}
    \hline
    \multirow{2}{*}{\bf Model}&
    \multicolumn{3}{c|}{\bf CoNLL-2014}\cr
    \cline{2-4}
    &P&R&$F_{0.5}$\cr
    \hline
    SG-GEC& 78.7&41.7&66.8\cr
    \hline
    $\text{SG-GEC}$  $-\text{Dependency relation prediction}$& 77.5&41.0&65.8\cr
     $\text{SG-GEC}$  $-\text{Distance prediction}$&78.3&41.5&66.5 \cr
   $\text{SG-GEC}$  $-\text{Ancestor-descendant relation prediction}$&78.1&41.2&66.2\cr
    
    \hline
    \end{tabular}
    \end{center}
    \caption{ Ablation results of sub-tasks for dependency tree correction on CoNLL-2014 benchmark.}
    \label{tab.abl2}
\end{table*}

\begin{table*}[h]
\begin{center}
\footnotesize
\begin{tabular}{|l| l|}
\hline
Standard Correction & 
$[\textbf { Do } \rightarrow \textbf { Does }]$ one who $[\textbf { suffered } \rightarrow \textbf { suffers }]$ from this disease keep it a secret ?\\
\hline  
Copy-Transformer  &    
$[\textbf { Do } \rightarrow \textbf { Does }]$ one who suffered from this disease keep it a secret ?\\
\hline 
SG-GEC  & 
$[\textbf { Do } \rightarrow \textbf { Does }]$ one who $[\textbf { suffered } \rightarrow \textbf { suffers }]$ from this disease keep it a secret ? \\
\hline
\end{tabular}
\end{center}
\caption{ Example of corrections.Brackets mark the spans of errors. The text on the right of arrow is the correction of the error on the left.}
\label{tab.case1}
\end{table*}

Firstly, we perform an ablation study on the CoNLL-2014 benchmark to evaluate
the influence of different modules in our proposed
SG-GEC model. Four ensemble models are investigated in this part: the baseline model Copy-Transformer, the model with dependency tree correction, the model with syntax-guided encoder, the full model SG-GEC. All models apply the data augmentation and techniques mentioned in previous parts. In order to ensure the reliability of the results, we do the significance tests of F-score between models. We randomly divide the test set into ten subsets and get the F-scores of the models on each of them. Then we apply paired-samples t-test to test the significance of results.

Table \ref{tab.abl} shows the results. As we can see, both graph attention mechanism and dependency tree correction task significantly improve the performance of model. We find that the additional dependency tree correction task can increase both precision and recall of the model. The model can achieve better performance by 
adding dependency tree correction loss. Besides, we can find that applying syntax-guided encoder with graph attention mechanism can greatly improve the performance of model as well. In particular, the precision of model with syntax-guided encoder increases by 5.2. It shows that graph attention mechanism can capture the syntactic knowledge within dependency tree and adopting this syntactic knowledge can significantly improve the precision of the model in grammatical error correction.

Utilizing both syntax-guided encoder and dependency tree correction task can further improve the model performance. 
By constructing the dependency trees of corrected sentences, we can enrich the model with the corrected syntactic knowledge. As we can see, our full model SG-GEC with both syntax-guided encoder and dependency tree correction task achieves the best precision, recall and F-score on the benchmark.

We further perform an ablation study to investigate the influence of proposed sub-tasks.  We evaluate the performance of the ensemble models that remove a sub-task on the CoNLL-2014 benchmark respectively. Table \ref{tab.abl2} presents the results.

We can find out that all sub-tasks, dependency relation prediction, distance prediction and ancestor-descendant relation prediction, are useful for improving the performance of the model. Specifically, the removing of dependency relation prediction sub-task has the greatest impact on the performance of the model. The F-score of the model without dependency relation prediction sub-task decreases by 1.0. It may indicate that the importance of dependency relations between the nodes. In addition, the F-score of the models without distance prediction and ancestor-descendant relation sub-task decreases by 0.3 and 0.6 respectively. We can see the effects of these two sub-tasks from the results.

\subsection{Case Study}

In this section, we use a specific case to analyze our proposed SG-GEC model.
In Table \ref{tab.case1}, we show an example of our proposed model.

\begin{figure}
  \centering
  \includegraphics[width=0.45\textwidth]{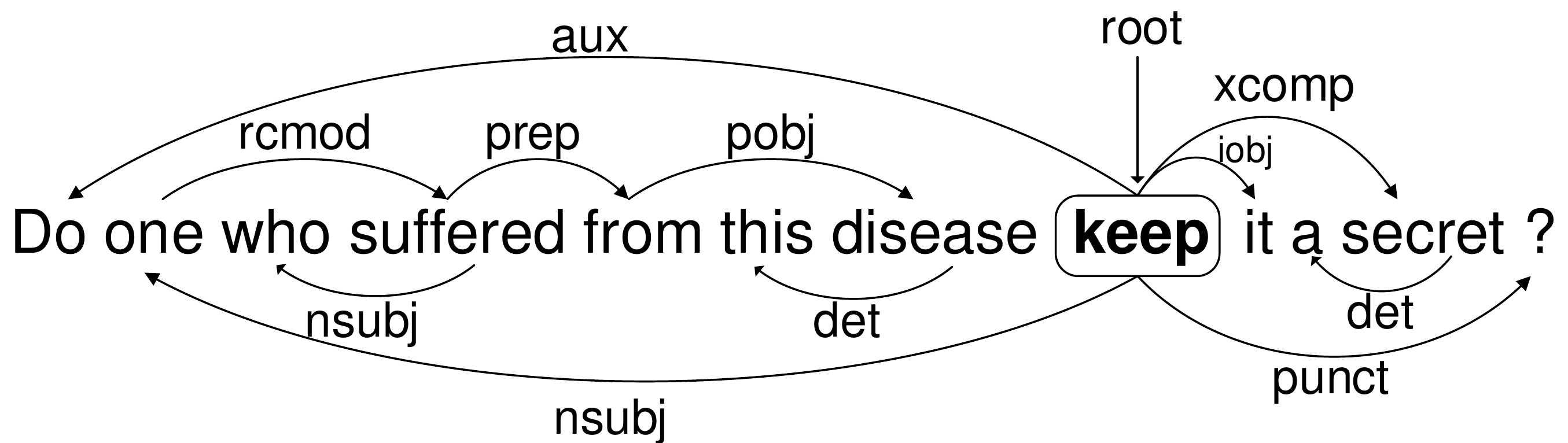} 
  \caption{Dependency tree of the source sentence in the case. } 
  \label{fig.tree1} 
\end{figure}

In this case, we can see how 
syntactic knowledge improves the performance of the model. The dependency tree of source sentence is shown as Figure \ref{fig.tree1}. The source sentence contains two grammatical errors: one is subject-verb agreement(SVA) error, and the other is verb tense(Vt) error. Copy-Transformer can only detect and correct the SVA errors. With the help of graph attention mechanism, SG-GEC model can capture the dependency relation rcmod(relative clause) between verb "suffered" and noun "one", and the dependency relation nsubj(nominal subject) between verb "keep" and noun "one". Considering the tense of verb "keep",  present tense is proper in this sentence. With the help of syntactic knowledge, our model can detect and correct both errors in this source sentence.



\section{Conclusion}

In this work, we propose a syntax-guided GEC model (SG-GEC) which adopts the graph attention mechanism to utilize the syntactic knowledge of dependency tree. Considering the dependency trees of the grammatically incorrect source sentences might provide incorrect syntactic knowledge, we propose a dependency tree correction task to deal with it. Combining with data augmentation method, our model achieves strong performances without using any large pre-trained models. We evaluate our model on public benchmarks of GEC task and it achieves competitive results.


\bibliographystyle{acl_natbib}

\end{document}